\documentclass[10pt,twocolumn,letterpaper]{article}

\usepackage{iccv}
\usepackage{times}
\usepackage{epsfig}
\usepackage{graphicx}
\usepackage{amsmath}
\usepackage{amssymb}

\usepackage[table]{xcolor}
\usepackage{booktabs}
\usepackage{float}
\usepackage{multicol}
\usepackage{appendix}

\usepackage[pagebackref=true,breaklinks=true,letterpaper=true,colorlinks,bookmarks=false]{hyperref}

 \iccvfinalcopy 

\def\iccvPaperID{2649} 

\begin{document}

\title{Consensus Feature Network for Scene Parsing}

\author{Tianyi Wu$^{1}$, Sheng Tang$^{2,}$\thanks{Corresponding author: Sheng Tang (ts@ict.ac.cn)}\,, Rui Zhang$^{2}$, Guodong Guo$^{1}$, Yongdong Zhang$^{2}$\\
$^1$Institute of Deep Learning, Baidu Research. \\
$^2$ Institute of Computing Technology, Chinese Academy of Sciences, Beijing, China. \\
{\tt\small {\{wutianyi01, guoguodong01\}@baidu.com, \{ts, zhangrui, zhyd\}@ict.ac.cn}}
}
\maketitle

\begin{abstract}
Scene parsing is challenging as it aims to assign one of the semantic categories to each pixel in scene images.
Thus, pixel-level features are desired for scene parsing.
However, classification networks are dominated by the discriminative portion, so directly applying classification networks to scene parsing will result in inconsistent parsing predictions within one instance and among instances of the same category.
To address this problem, we propose two transform units to learn pixel-level consensus features.
One is an Instance Consensus Transform (ICT) unit to learn the instance-level consensus features by aggregating features within the same instance.  
The other is a Category Consensus Transform (CCT) unit to pursue category-level consensus features through keeping the consensus of features among instances of the same category in scene images.  
The proposed ICT and CCT units are lightweight, data-driven and end-to-end trainable.
The features learned by the two units are more coherent in both instance-level and category-level.
Furthermore, we present the Consensus Feature Network (CFNet) based on the proposed ICT and CCT units, and demonstrate the effectiveness of each component in our method by performing extensive ablation experiments. Finally, our proposed CFNet achieves competitive performance on four datasets, including Cityscapes, Pascal Context, CamVid, and COCO Stuff. 

\end{abstract}

\section{Introduction}

\begin{figure}[t]
\begin{center}
   \includegraphics[width=\linewidth]{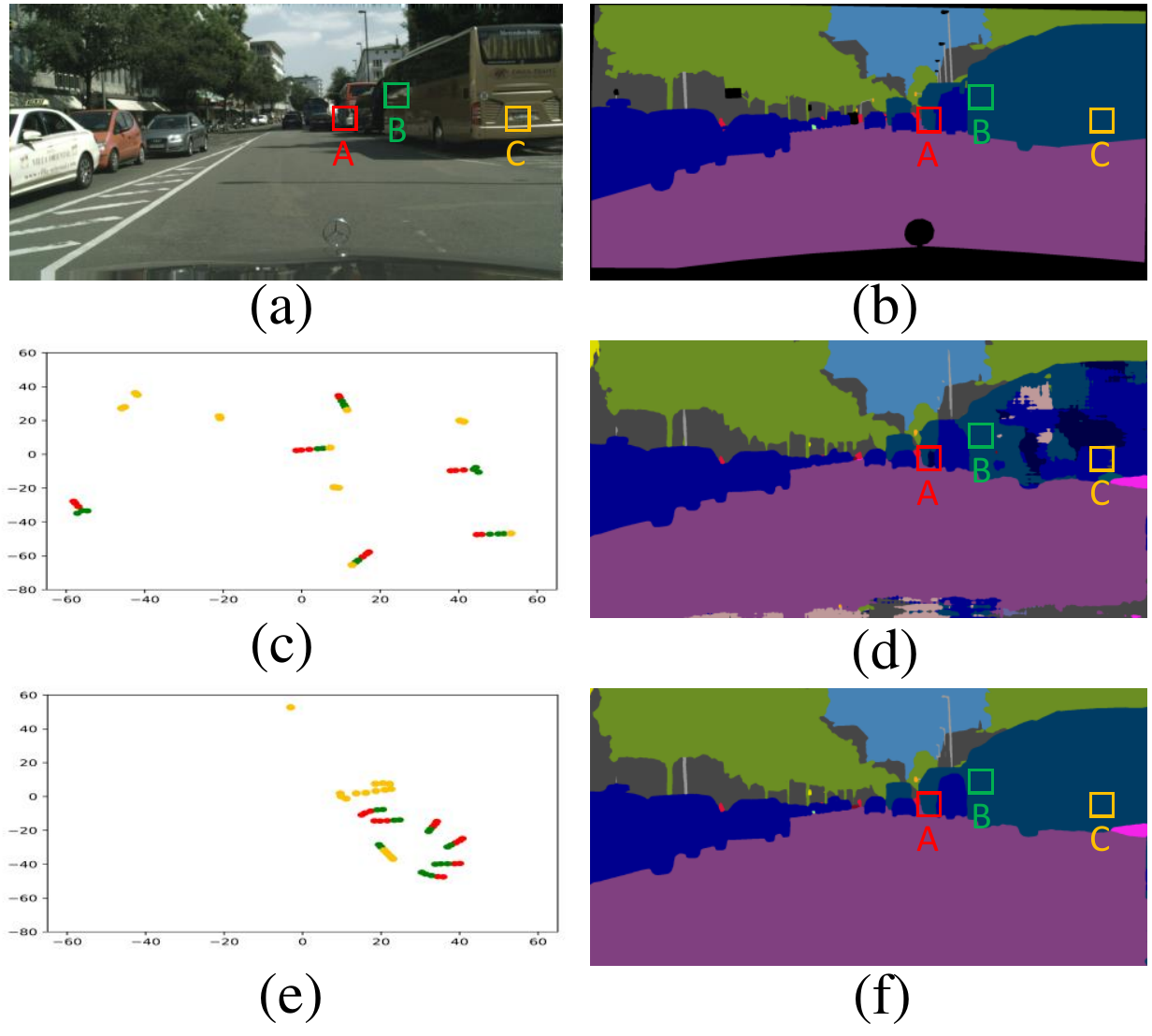}
\end{center}
   \caption{Visualization using t-SNE \cite{maaten2008visualizing} to illustrate features learned from FCN (ResNet-101) and the proposed CFNet. (a) Input image, in which region B and C are within the same instance, and region A and C belong to the same category. (b) Ground truth. (c) Features of A, B and C learned by FCN are far apart. (d) FCN has inconsistency parsing predictions within one instance and those of the same category. (e) Features of A, B and C learned by the proposed CFNet are coherent and indistinguishable. (f) The proposed CFNet has consistent parsing predictions within instances of the same category. (Best viewed in color)
   }
\label{fig:fig1}
\vspace{-15pt}
\end{figure}

Scene parsing has been an essential component for scene understanding and can play a crucial role in applications such as auto-driving, auto-navigation, and virtual reality. The goal of scene parsing is to label each pixel to one of the semantic categories including not only discrete objects (e.g., car, bicycle, people) but also stuff (e.g., road, sky, bench).

Deep Convolutional Neural Networks (DCNNs) have achieved remarkable progress in semantic segmentation or scene parsing. Currently, most of the successful methods for scene parsing are based on classification networks \cite{simonyan2014very, he2016deep, huang2017densely}. However, there are some limitations for taking classification networks as the feature extractor for scene parsing. Classification networks tend to learn the image-level representation of the whole input examples. Moreover, previous works \cite{zeiler2014visualizing,zhang2018adversarial,hou2018self} show that the image-level representation is often dominated by the discriminative portion of the foreground or predominant objects, e.g., the horse's head and dog's face.
However, scene parsing aims to parse both discrete objects and stuff things, so the pixel-level features are desired. 
Therefore, directly applying classification networks to scene parsing will result in two drawbacks, as shown in Fig. ~\ref{fig:fig1}(d): (1) The intra-class features of all spatial positions of dominated objects are not consistent, leading to inconsistent parsing predictions within one instance.  (2) The inter-class features of non-discriminative regions (e.g., subordinate objects and stuff) are easily confused, resulting in the inconsistent prediction of instances of the same category.

To address the above problem, we expect to learn the pixel-level consensus features for scene parsing.
The consensus features are inspired by neighborhood consensus \cite{stockman1982matching, zhang1995robust,schmid1997local,chang1997fast, rocco2018end, Rocco18b} which finds reliable dense correspondences between a pair of images in object matching. In this work, we aim to learn the consensus features which are indistinguishable for pixels within an instance or a category.
The consensus features contain two aspects: instance-level and category-level. 
As shown in Fig.~\ref{fig:fig1} (a), (1) features of regions in the same instance (e.g., B and C) should keep the instance-level consensus, and (2) features of regions in different instances with the same category (e.g., A and C) should maintain the category-level consensus.

To learn the consensus features, we propose two consensus transform units, including Instance Consensus Transform (ICT) unit and Category Consensus Transform (CCT) unit. The ICT unit is expected to learn the instance-level consensus features. Specifically, we introduce a tiny local network (abbreviated as LN) to generate the instance-level transform parameters for each pixel by using surrounding contextual information. Then we apply the instance-level transform parameters to aggregate features within the same instance. 
On the other hand, due to multiple instances of the same category in the scene images,
we employ the CCT unit to pursue the category-level consensus features. Specifically, we introduce a lightweight global network (abbreviated as GN) to generate the category-level transform parameters. Different from LN, GN aims to model the interaction at specific locations with respect to all other locations. 
The proposed two units are learned in a data-driven manner without any extra supervision.
We update features at all positions with these two units. 
For each position, the two units can adaptively strengthen the information of relative locations (regarded as foreground) and suppress the irrelative locations (regarded as background).
Thus, the consensus features are indistinguishable within the foreground and invariant to the background variations.
Compared with FCN in Fig.\ref{fig:fig1}(c), The features learned by the two units are more coherent in instance-level and category-level, as shown in Fig.\ref{fig:fig1}(e). 
Meanwhile, the inconsistent parsing prediction in Fig.\ref{fig:fig1}(d) is corrected by the proposed methods, as shown in Fig.\ref{fig:fig1}(f).

Based on the proposed ICT and CCT units, we present a new scene parsing framework, called Consensus Feature Network (CFNet), to learn pixel-level consensus features and obtain consistent parsing results. 
We demonstrate the effectiveness of each component in our proposed approach by performing extensive ablation experiments. Furthermore, our model achieve competitive performance on four challenging scene parsing datasets, Cityscapes \cite{cordts2016cityscapes}, PASCAL Context \cite{mottaghi2014role}, CamVid \cite{brostow2008segmentation} and COCO Stuff \cite{Caesar_2018_CVPR}.


\section{Related Work}

In 2015, Long \etal proposed FCN \cite{long2015fully}, which is the first approach to adopt classification networks to get dense output and end-to-end training. Later, how to better adjust the classification network for scene parsing has attracted more and more attention. Hence, we review several aspects of research related to this work.

Contextual information plays a vital role in scene understanding \cite{belongie2002shape,tu2005image}. Recent works \cite{chen2017rethinking, Ding_2018_CVPR, Yu_2018_CVPR, zhang2018context, Fu_2019_ICCV} have shown that contextual information is helpful for models to make a better local decision. One direction is to append context aggregation modules to learn contextual information. Liu \etal proposed ParseNet \cite{liu2015parsenet}, which uses global context to augment the feature at each location. Chen \etal \cite{chen2016deeplab} introduced atrous spatial pyramid pooling to learn contextual information. However, Zhao \etal \cite{Zhao_2017_CVPR} introduced a pyramid pooling module to exploit global information from different subregions. Zhang \cite{zhang2017global} proposed to use context to refine the inconsistent parsing results iteratively. More recently, Ding \etal \cite{Ding_2018_CVPR} proposed a novel context contrasted local feature that not only leverages the informational context but also spotlights the local information in contrast to the context. Zhang \etal \cite{zhang2018context} introduced a Context Encoding Module which can capture global context and selectively highlight the class-dependent feature maps. 
Zhao \etal \cite{Zhao_2018_ECCV} proposed to relax the local neighborhood constraint for enhancing information flow.
However, very recent works \cite{fu2018dual, yuan2018ocnet, Huang_2019_ICCV, Zhu_2019_ICCV, Li_2019_ICCV} employ the self-attention mechanism  to learn the global interdependences of features. In contrast to them, we propose to exploit surrounding contextual information and long-range dependencies to generate the parameters of consensus transforms.  


\begin{figure*}[t]
\begin{center}
   \includegraphics[width=0.95\linewidth]{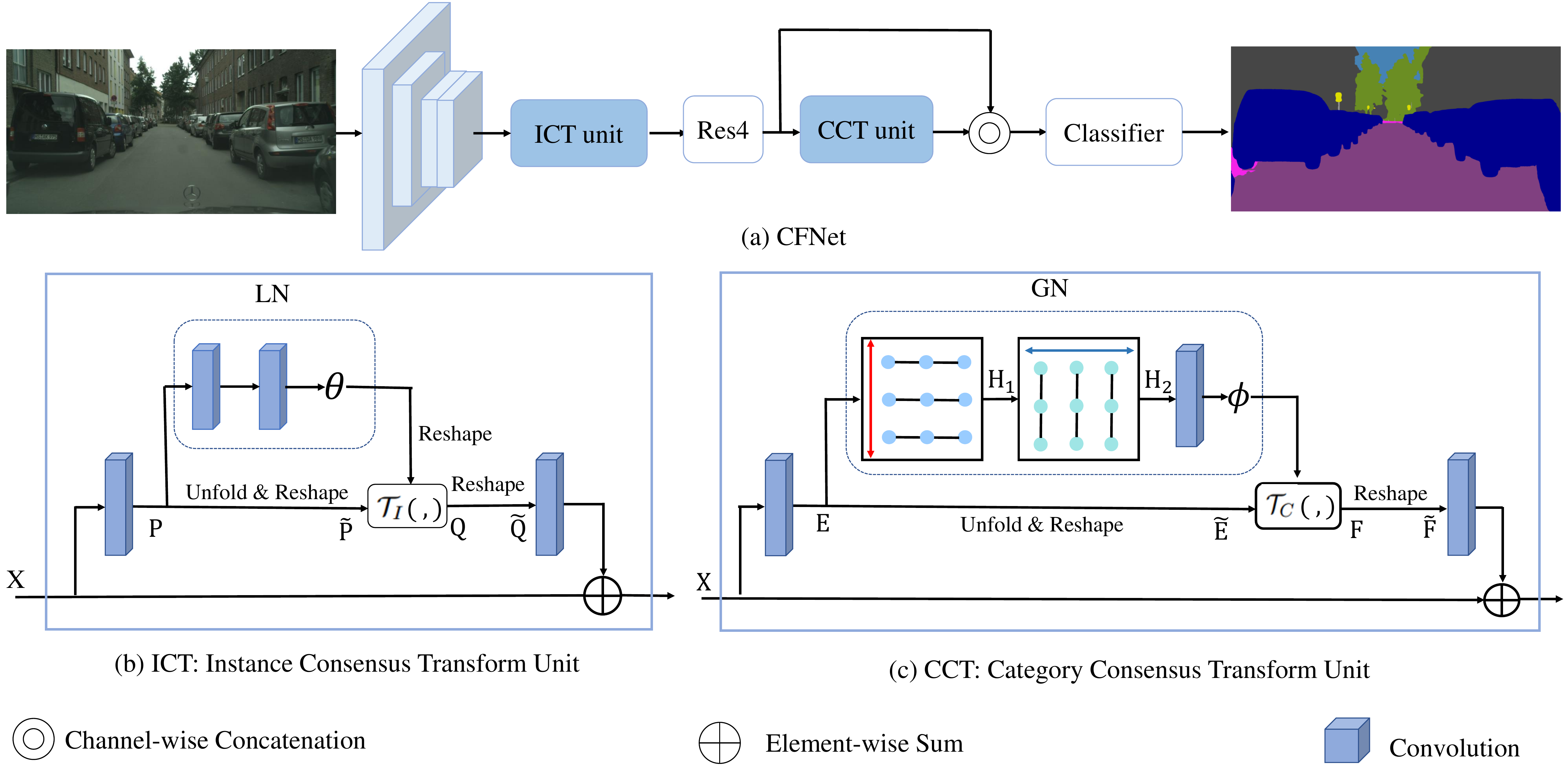}
\end{center}
   \caption{An overview of the Consensus Feature Network (CFNet). (a) Network Architecture. We take ResNet-101 as the backbone, the ICT and CCT units are inserted to ResNet-101 on Res3 and Res4 for learning the consensus feature. (b) Components of the Instance Consensus Transform (ICT) unit. (c) Components of the Category Consensus Transform (CCT) unit. The ICT and CCT units are applied to pursuit the instance-level and category-level consensus respectively. Residual connection is employed in the ICT and CCT units, which improves gradient propagation.
}
\label{fig:fig2}
\vspace{-15pt}
\end{figure*}
Another relevant aspect of related works is how to suppress responses from the background. The existing works have been concerned with handcrafted features for achieving a similar property. For example, Trulls  \cite{trulls2013dense} proposed an embedding method where the Euclidean distance measures how likely it is that two pixels will belong to the same region. Harley \etal \cite{harley2015learning} designed an embedding space for estimating the pair-wise semantic similarity and used a contrastive side loss to train the ``embedding" branch. Following this, the segmentation-aware convolution \cite{harley2017segmentation} is proposed to attend to inputs according to local masks. In these works, the embeddings are defined in a handcrafted manner, or a specific loss function is required to guide the process of training. Instead, we use one neural network to learn desired transforms automatically without adding any other supervision, and the learned transforms are adaptive to test examples.


Most closely related to our work are the neighborhood consensus \cite{bian2017gms, Rocco18b} and PiCANet \cite{liu2018picanet}. The locally consistent matche \cite{bian2017gms} was proposed for measuring neighborhood consensus. Rocco \cite{Rocco18b} developed a neighborhood consensus network to learn neighborhood consensus constraints, which analyze the full set of dense matches between a pair of image and learns patterns of locally consistent correspondences. PiCANet  \cite{liu2018picanet} proposed a novel attention network for learning global contrast and homogeneousness. In contrast to them.
we propose the consensus transforms, which analyzes the pixel-wise feature matches and transforms in each instance or instances of the same category. 


\section{Approach}

In this section, we present the details of the proposed Consensus Feature Network (CFNet) for scene parsing. First, we will introduce the general framework of the proposed method. Then, we will present the ICT and CCT units which are employed to achieve instance-level and category-level consensus, respectively.

\subsection{Overview}

The network architecture is illustrated in Fig.~\ref{fig:fig2}(a). An input image is fed into a classification network (ResNet-101) pre-trained on ImageNet, which is adapted to a fully convolutional fashion \cite{long2015fully}. Similar to previous works \cite{Zhao_2017_CVPR, zhang2018context}, dilated convolutions are employed in Res3 and Res4 with the output size of $1/8$. Previous works \cite{zeiler2014visualizing, Yu_2018_CVPR} have shown that the network encodes finer spatial information in the lower stages, and learn richer semantic feature in the higher stages. 
Therefore, we choose semantic-level features to conduct the consensus transforms. 
To perform the consensus transforms friendly, we need to get the instance-level consensus before reaching category-level consensus. So the ICT and CCT units are added after Res3 and Res4, respectively. 
Then we predict the label for each pixel according to the transformed feature maps, and up-samples the label map for $8\times$  times at last.
The proposed ICT and CCT units will be described and formulated in detail as follows.

\begin{figure}[t]
\begin{center}
   \includegraphics[width=\linewidth]{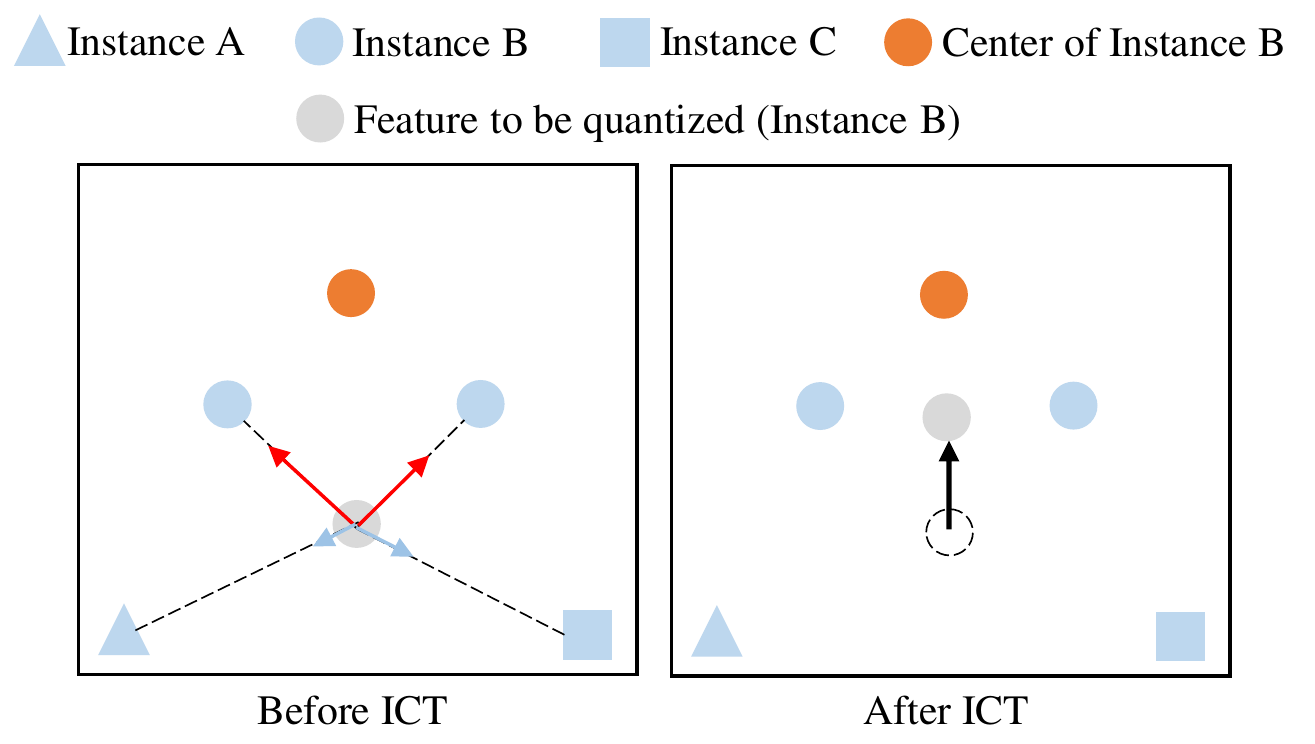}
\end{center}
   \caption{Illustration of the Instance Consensus Transform. A response can be reconstructed by the features at the surrounding window. 
   The length of the arrow in the left subfigure indicates the interaction intensity.
   }
\label{fig:fig3}
\vspace{-10pt}
\end{figure}

\subsection{Instance Consensus Transform Unit}

To achieve the instance-level consensus, we propose the ICT unit to learn the instance-level consensus features. We do not employ an object detector pre-trained on an additional dataset to find each object, but approximate this process by conducting the transform in a surrounding window (for a given position). Fig.~\ref{fig:fig3} illustrates this transform, given a target point, its responses can be reconstructed by features at the surrounding window. Specifically, features belonging to the same instance are enhanced, features of other instance are weakened. From the right subfigure of Fig.~\ref{fig:fig3}, we can observe that the transformed feature is more coherent. 

The ICT unit uses surrounding contextual information to generate the transform parameters for each spatial location. As shown in Fig.~\ref{fig:fig2}(b), for a local feature map $X \in \mathbb{R}^{C \times H \times W }$, with $C$ being the feature dimension and $H \times W $ the spatial size, the ICT unit firstly applies one convolution layers with $1 \times 1$ filters on $X$ to reduce dimension for saving computation, while obtaining the feature map $P$, where $P \in \mathbb{R}^{ C_1 \times H \times W }$. $C_1$ is the channel number of feature maps, which is less than $C $ (typically $C_1= \frac{C}{4}$).

After obtaining the feature map $P$, the ICT unit further employs a lightweight local network (LN) to generate the parameters of the ICT $ \theta \in \mathbb{R}^{H \times W \times r^2}$, where $r$ represents the size of the local region centered on the current spatial position (typically $r=5$). The size of $ \theta $ can vary depending on the local region size $r$. We expect LN to generate transform parameters for each pixel by using the corresponding surrounding context information. We instantiate LN with two convolutional layers, whose filter size is $r \times r$ and $1 \times 1$, respectively. The first convolutional layer ($r \times r$) is employed to capture surrounding contextual information, and then is fed into the second convolutional layer ($1 \times 1)$ for generating the parameters of the ICT. Then the transform parameters $ \theta $ are reshaped into $\tilde{ \theta} \in \mathbb{R}^{N \times r^2}$, where $N= H \times W$. Meanwhile, the feature map $P$ is conducted with an unfold operation for extracting sliding local feature blocks and is reshaped for obtaining the feature map $ \tilde{P} \in \mathbb{R}^{ C_1 \times N \times r^2}$. We define function $ \mathcal{T}_{I}(x,y)$ as the elemental multiplication of the tensor $x$ and $y$, and then sum it according to the last dimension. The new feature map $Q \in \mathbb{R}^{ C_1 \times N }$ are generated by
\begin{align}
Q= \mathcal{T}_{I}( \tilde{P}, \tilde{\theta}).
\label{eq1}
\end{align}
 Next, we reshape $Q$ to $ \tilde{Q} \in \mathbb{R}^{C_1 \times H \times W}$.
In particular, any feature vector $ \tilde{Q}_{ij} \in \mathbb{R}^{C_1} $ in $ \tilde{Q} $ at position $(i,j)$ is multiplication of the associated neighbours $(h,w) \in \mathbb{N} (i,j) $ in $P$ and the corresponding instance-level consensus transform parameters $ \theta_{ij} \in \mathbb{R}^{r^2}$, where $i \in [1, H], j \in [1, W]$, $ \mathbb{N}(i,j)$ is a $r \times r$ square with the center of $(i,j)$. So the transform operator at each location $(i,j)$ can be formulated as:
\vspace{-10pt}
\begin{align}
\tilde{Q}_{ij}= \frac{1}{r^2}\sum_{h,w \in \mathbb{N} (i,j)} \theta_{ij}(h,w)P_{hw},
\label{eq2}
\end{align}
where $\theta_{ij}(h,w)= \theta_{ij}(hr+w+ \frac{r^2-1}{2})$.
Eqn.~\eqref{eq2} encapsulates the transform of various handcrafted filters in a generalized way. For the bilateral filter \cite{tomasi1998bilateral}, $\theta_{ij}(h,w)$ is a Gaussian that jointly captures RGB and geometric distance between pixels $(i,j)$ and $(h,w)$. For the mean filter \cite{schalkoff1989digital},  $\theta_{ij}(h,w)= 1$.

After obtaining the feature map $\tilde{F}$, we apply one convolution layer with $1 \times 1$ filters for dimension expansion, so that the output dimension can match the dimension of input $X$, forming a residual connection.

\subsection{Category Consensus Transform Unit}
It is very useful for high-quality scene segmentation to achieve category-level consensus, since there are usually multiple objects of the same class in the scene images. For example, for Cityscapes \cite{cordts2016cityscapes} dataset, there are 7 humans and 14 vehicles per image on averages. Therefore, we propose the Category Consensus Transform (CCT) unit to pursue category-level consensus features.


The structure of CCT unit is illustrated in Fig.~\ref{fig:fig2}(c), we deploy a global network (GN) to generate the category-level consensus transform parameters $\phi \in \mathbb{R}^{N \times N}$, where $N= H\times W$. We expect GN to have the ability of ``seeing" the whole feature map $E$, and has the ability to model the interaction between each location and other locations across the whole input feature maps. A natural solution is to employ a fully connected layer, global convolution, stacked multiple large kernel convolutions. These solutions are not very effective, since they introduce a huge number of parameters or memory usage.

Inspired by \cite{visin2015renet} which introduce recurrent neural networks to model region-wise dependency, we instantiate GN with two bidirectional LSTMs (BiLSTMs) and one convolutional layer with $1 \times 1$ filters. We use the first BiLSTM to scan the feature maps in bottom-up and top-down directions, as shown in Fig.~\ref{fig:fig2}(c). It takes a row-wise feature as input for one time and updates its hidden state. A typical LSTM unit contains an input gate $i_t$, a forget gate $f_t$, an output gate $o_t$, an output state $h_{t}^{v}$, and an internal memory cell state $c_t$. The rule of scanning $\mathcal{H}_{v}$ can be formulated as follows:
\begin{align}
h_{t}^{v} = \mathcal{H}_{v}(h_{t-1}^{v}, E_{t}^{v}).
\label{eq3}
\end{align}
The detailed computation is described as follows:
\begin{align}
 \begin{pmatrix} i_t \\ f_t \\ o_t \\ u_t  \end{pmatrix} = \begin{pmatrix} \sigma \\ \sigma \\ \sigma \\ tanh \end{pmatrix}
 \begin{pmatrix} W \begin{pmatrix} E_{t}^{v} \\ h_{t-1}^{v} \end{pmatrix} \end{pmatrix},
\label{eq4}
\end{align}
\vspace{-15pt}
\begin{align}
c_t = f_t \odot c_{t-1} + i_t \odot u_t ,
\label{eg5}
\end{align}
\vspace{-25pt}
\begin{align}
h_{t}^{v} = o_t \odot tanh( c_{t-1} ) ,
\label{eg6}
\end{align}
where $\odot$ means element-wise product, $t \in [1, H]$, $E_{t}^{v} \in \mathbb{R}^{C \times W}$ indicates the sliced feature map and is the input to the LSTM at time step t, and $u_t$ denotes the modulated input. $\sigma$ is the sigmoid activation function. $W: \mathbb{R}^{(C+d) \times W} \rightarrow \mathbb{R}^{4d \times W}$ is an affine transform consisting of the parameters of LSTM, where $d$ is the number of LSTM cell state units. However, the BiLSTM computes the forward hidden sequence $\overrightarrow{h^{v}}$ and the backward hidden sequence $\overleftarrow{h^{v}} $ by iterating the forward layer form $t=1$ to $H$, the backward layer from $t=H$ to $1$ simultaneously. The calculations of BiLSTM can be formulated as follows:
\begin{align}
\overrightarrow{h_{t}^{v}} = \mathcal{H}_v (\overrightarrow{h_{t-1}^{v}}, E_{t}^{v}),
\label{eg7}
\end{align}
\vspace{-25pt}
\begin{align}
\overleftarrow{h_{t}^{v}} = \mathcal{H}_v (\overleftarrow{h_{t+1}^{v}}, E_{t}^{v}).
\label{eg8}
\end{align}
After bi-direction sweeping, we concatenate the hidden states $\overrightarrow{h_{t}^{v}}$ and  $\overleftarrow{h_{t}^{v}}$ to get a composite feature map $H_1$. In a similar manner, we employ the second BiLSTM to sweep over feature maps $H_1$ horizontally, which takes a column-wise feature slice as input for one time and updates its hidden state. Then we concatenate the forward and backward hidden states for the second BiLSTM to get feature maps $H_2$, which is taken as the representation of global interaction between each spatial position and all other locations.
Then each response in feature maps $H_2$ is an activation at the specific location with respect to the whole image. 
Afterward, the global interaction information is fed into the $1\times 1$ Conv layer to generate the transform parameters $\phi$. We define function $\mathcal{T}_{C}(x, y)$ as the matrix product of tensor $x$ and tensor $y$. The new feature maps $F $ is generated by
\begin{align}
F= \mathcal{T}_{C}(\tilde{E}, \phi)= \tilde{E}\phi,
\label{eq9}
\end{align}
where $F \in \mathbb{R}^{C_1 \times N}$. Next, we reshape $F$ to feature maps $ \tilde{F} \in \mathbb{R}^{C_1 \times H \times W}$.
In pariticular, any feature vector $\tilde{F}_{ij} \in \mathbb{R}^{C_1}$ in $F$ at position $(i,j)$ is generated by
\begin{align}
\tilde{F}_{ij} = \frac{1}{H\times W}\sum_{h=1,w=1}^{H,W} \phi_{ij}(h,w)E_{hw},
\label{eq10}
\end{align}
where $i \in [1, H], j \in [1, W]$, $\phi_{ij} \in \mathbb{R}^{N}$, $\phi_{ij}(h,w) = \phi_{ij}(hw)$, $E_{hw} \in \mathbb{R}^{C_1}$ is the feature at the location $(h,w) $ on feature maps $E$.

Note that if $\phi_{ij}(h,w)$ learns the responses based on the relationship between $E_{ij}$ and $E_{hw} $, Eqn.~\eqref{eq10} is equivalent to non-local operation \cite{wang2018non}. The non-local unit generates an attention map for the feature which has a limited receptive field. For the CCT unit, the response between any two points is not simply a matter of modeling the relationship between two features, but the interaction of other features with them. Furthermore, the response in the non-local unit is computed by handcrafted pairwise function (e.g., gaussian, embedded gaussian, dot product), but our parameters are dynamically generated by the GN, which is adaptive to each test example.

After obtaining global consensus feature maps $\tilde{F}$, we apply one convolution layer with $1 \times 1$ filters for dimension expansion. Finally, residual learning is employed to improve the gradient back-propagation during training.

\section{Experiments}
To validate the proposed approach, we conduct comprehensive experiments on multiple datasets, including Cityscapes dataset \cite{cordts2016cityscapes}, PASCAL Context dataset \cite{mottaghi2014role}, CamVid dataset \cite{brostow2008segmentation},  and COCO Stuff dataset \cite{Caesar_2018_CVPR}.
In the following subsections, we first describe the datasets and the experimental settings. Then the contributions of each component are investigated in ablation experiments on the Cityscapes dataset. Finally, we report our results on four scene parsing benchmarks, i.e., Cityscapes dataset, PASCAL Context, CamVid, COCO Stuff, and compare our proposed approach with the state-of-the-art approaches. 

\subsection{Datasets}

\noindent\textbf{Cityscapes Dataset}\quad
The dataset contains 5, 000 finely annotated images and 20, 000 coarsely annotated images collected in street scenes from 50 different cities, which is targeted for urban scene segmentation. Only the 5, 000 finely annotated images are used in our experiments, divided into three subsets, including 2, 975 images in training set, 500 images in validation set and 1, 525 images in test set. High-quality pixel-level annotations of 19 semantic classes are provided in this dataset.

\noindent\textbf{PASCAL Context Dataset}\quad
The dataset involves 4, 998 images in training set and 5, 105 images in the test set. It provides detailed semantic labels for the whole scene. Similar to \cite{wu2018tree, zhang2018context}, the proposed approach is evaluated on the most frequent 59 categories and 1 background class.

\begin{table}[ht]
\begin{center}
\begin{tabular}{|p{4cm}|p{2cm}<{\centering}|}
\hline
Method                 & mIoU (\%) \\
\hline\hline
Res101 (baseline)            &74.9   \\
\hline
Res101 + ICT (0,5)           &74.3    \\
Res101 + ICT (1,5)           &75.3    \\
Res101 + ICT (2,5)           &75.7    \\
Res101 + ICT (3,5)           &78.8    \\
Res101 + ICT (4,5)           &78.6    \\
\hline
\end{tabular}
\end{center}
\caption{Ablation experiments of ICT on the validation set of Cityscapes. ICT represents the Instance Consensus Transform unit. Without loss of generality, ``ICT(3,5)'' means the ICT unit with $r=5$ is inserted to ResNet-101 on Res3.}\label{table:tab1}
\end{table}

\begin{table}[ht]
\begin{center}

\begin{tabular}{|p{4cm}|p{2cm}<{\centering}|}
\hline
Method             & mIoU (\%) \\
\hline\hline
Res101 (baseline)               &74.9\\
\hline
Res101  + ICT ($r=3$)           & 77.4   \\
Res101  + ICT ($r=5$)           & 78.8   \\
Res101  + ICT ($r=7$)          &  76.8  \\
\hline
\end{tabular}
\end{center}
\caption{Ablation experiments of ``r" on the validation set of Cityscapes. ``$r$" indicates the size of the local window in the ICT unit. }
\label{table:tab2}
\end{table}
\noindent\textbf{CamVid Dataset}\quad
The CamVid is a road scene dataset from the perspective of a driving automobile. The dataset involves 367 training images, 101 validation images, and 233 test images. The images have a resolution of 960 $\times$ 720. Following \cite{kundu2016feature, badrinarayanan2017segnet, amirul2017gated, bilinski2018dense}, we consider 11 larger semantic classes (road, building, sky, tree, sidewalk, car, column-pole, fence, pedestrian, bicyclist, and sign-symbol) for evaluation.

\noindent\textbf{COCO Stuff Dataset}\quad
The dataset contains 10, 000 images from Microsoft COCO dataset \cite{lin2014microsoft}, out of which 9, 000 images are for training and 1, 000 images for testing. The unlabeled stuff pixels in original images of Microsoft COCO are further densely annotated with extra 91 classes. Following \cite{Ding_2018_CVPR}, we evaluate the proposed method on 171 semantic classes including 80 objects and 91 stuff annotated to each pixel.

\subsection{Experimentation Details}

We take ResNet-101 \cite{he2016deep} pre-trained on ImageNet as the backbone. Similar previous works \cite{Zhao_2017_CVPR, zhang2018context}, dilated convolutions are employed in Res3 and Res4 with the output size of $1/8$. The output predictions are upsampled 8 times using bilinear interpolation. Meanwhile, we replace the standard Batchnorm with InPlace-ABN \cite{rota2018place} to the mean and standard-deviation of BatchNorm across multiple GPUs. The SGD with mini-batch is used for training. Following prior work \cite{Zhao_2017_CVPR,zhang2018context}, we use the ``poly'' learning rate policy, where the learning rate is multiplied by $  (1-\frac{iter}{total\_iter})^{power}$ with $power = 0.9$. The base learning rate is set to 0.01 for Cityscapes. The momentum is set to 0.9 and weight decay is set to 0.0001. For data augmentation, we adopt randomly
scaling in the range of [0.5,2] and then randomly cropping the image into a fixed size using zero-padding if necessary. For loss function, we employ the cross-entropy loss on both

\begin{table}[ht]
\begin{center}

\begin{tabular}{|p{4.5cm}|p{2cm}<{\centering}|}

\hline
Method          & mIoU (\%) \\
\hline\hline
Res101 (baseline)                   & 74.9          \\
\hline
Res101 + CCT (1x1 Conv)             & 75.7          \\
Res101 + CCT (Global Conv)          & 76.9          \\
Res101 + CCT (Large Kernel)          & 75.3          \\
Res101 + CCT (Stacked Conv)          & 76.7          \\
Res101 + CCT (BiLSTM)               & 77.5          \\
\hline
\end{tabular}
\end{center}
\caption{Comparison of different instantiations of CCT unit on the validation set of Cityscapes. CCT represents the Category Consensus Transform unit.}
\label{table:tab3}
\vspace{-10pt}
\end{table}


\begin{table}[ht]
\begin{center}
\begin{tabular}{|p{4.5cm}|p{2cm}<{\centering}|}
\hline
Method          & mIoU (\%) \\
\hline\hline
Res101 (baseline)             & 74.9          \\
\hline
Res101 + NL                   &  76.8         \\
Res101 + CCT (BiLSTM)         &  77.5         \\
\hline
\end{tabular}
\end{center}
\caption{Comparison of Non-local and CCT unit  on the validation set of Cityscapes. ``NL" indicates non-local unit \cite{wang2018non}. }
\label{table:tab4}
\end{table}

\begin{table}[ht]
\begin{center}
\begin{tabular}{|p{3cm}|p{2cm}<{\centering}|p{2cm}<{\centering}|}
\hline
Method         & mIoU (\%)     & Times\\
\hline\hline
Res101 (baseline)                   & 74.9     & 103.8 ms     \\
\hline
Res101 + ICT                        & 78.8        & +4.8 ms   \\
Res101 + CCT                        & 77.5         & +17.1 ms  \\
Res101 + ICT + CCT                  & 79.9       & +22.3 ms   \\
\hline
\end{tabular}
\end{center}
\caption{Ablation experiments of ICT and CCT on Cityscapes validation set. }
\label{table:tab5}
\vspace{-10pt}
\end{table}

\noindent the final output of CFNet and intermediate output from `Res3'. Similar to the original setting introduced by Zhao \etal \cite{Zhao_2017_CVPR}, the weight over the main loss and auxiliary loss is set to 1 and 0.4 respectively. The performance is reported using the commonly mean Intersection-over-Union (IoU). We use a single-scale evaluation to compute mean IoU in all ablation experiments. For evaluation, we average the network prediction in multiple scales following\cite{chen2016deeplab, Zhao_2017_CVPR, zhang2018context}.

\subsection{Experiments on Cityscapes}

\subsubsection{Which stage to add ICT unit?}
Tab.~\ref{table:tab1} compares the proposed ICT unit added to different stages of ResNet. The unit is added after the last residual block of a stage. We use ``(n,r)" to indicate insert location and the region size of the Instance Consensus Transform. For example, ``Res101 + ICT (1,5)" means the ICT unit with $r=5$ is inserted to ResNet-101 on Res1. As shown in Tab.~\ref{table:tab1}, the improvement over the baseline of the ICT unit on Res3 and Res4 is similarly significant, while on the Res1 and Res2 are slightly small. It is also interesting to see that the ``Res101 + ICT (0,5)" method achieves slight lower mIoU than the baseline (74.3 vs. 74.9).  Our conjecture is that the consensus transforms require features with semantic-level information, yet the lower stage of the network tends to learn spatial-level information. For subse- 

\begin{table}[ht]
\begin{center}
\begin{tabular}{|p{2.8cm}p{2.2cm}<{\centering}p{1.5cm}<{\centering}|}

\hline
Method  & Backbone & mIoU (\%) \\
\hline\hline
DeepLab-v2 \cite{chen2016deeplab}       &ResNet-101   &70.4\\
RefineNet \cite{lin2017refinenet}      &ResNet-101   &73.6\\
SAC \cite{zhang2017scale}              &ResNet-101   &78.1\\
PSPNet \cite{Zhao_2017_CVPR}         &ResNet-101   &78.4\\
BiSENet \cite{yu2018bisenet}        &ResNet-101      &78.9\\
AAF \cite{Ke_2018_ECCV}              &ResNet-101      &79.1 \\
DFN \cite{Yu_2018_CVPR}              &ResNet-101      &79.3\\
TKCN \cite{wu2018tree}              &ResNet-101      &79.5\\
PSANet \cite{Zhao_2018_ECCV}        &ResNet-101      &80.1 \\
DenseASPP \cite{yang2018denseaspp}    &DensetNet-161   &80.6\\
GloRe \cite{chen2019graph}            &ResNet-101     & 80.9 \\
SPGNet \cite{Cheng_2019_ICCV}   & $2 \times$ ResNet-50  & 81.1 \\
CCNet \cite{Huang_2019_ICCV} $\dagger$     & ResNet-101               & 81.4 \\
\hline
CFNet (ours)                              &ResNet-101      & 81.3\\
\hline
\end{tabular}
\end{center}
\caption{Scene parsing results on Cityscapes test set. All results are evaluated by the official evaluation server. Our method only train on both train-fine and val-fine set, without using extra ``coarse" training set.
$\dagger$ indicates employing OHEM.}
\label{table:tab6}
\vspace{-15pt}
\end{table}
\noindent quent  experiments, we fix the ICT unit behind Res3, which has 3.9\% improvement over the baseline.


\subsubsection{Different sizes of $r$ in ICT unit}
Tab.~\ref{table:tab2} compares different sizes of $r$ when the ICT unit is added after Res3. From the experimental results in Tab.~\ref{table:tab2}, we can see that increasing $r$ (from 3 to 5) can improve performance, however, performance will drop obviously when increasing $r=7$, which shows that choosing the right $r$ is important for the instance-level consensus transform. For subsequent experiments, we configure ICT with $r=5$, which has 3.9\% improvement over the baseline ( 78.8\% vs. 74.9\%).


\subsubsection{Different instantiations of CCT unit}
Tab.~\ref{table:tab3} compares different types of GN in the CCT unit. (1) $1 \times 1$ Conv: we simply instantiate GN with a $1 \times 1$ convolution layer, which has a limited receptive field (relative to input feature maps), and is employed for generating the parameters of CCT. (2) Global Conv: we use one $33 \times 33$ convolution with dilation = 3, which takes whole input feature maps as the receptive field. (3) Large Kernel \cite{peng2017large}, which combine $1\times N$ and $N \times 1$ convolution. (4) Stacked Conv: It means that two $24 \times 24 $ convolution with dilation = 2 is used. (5) Bidirectional LSTM: GN is instantiated with two BiLSTMs and a $1 \times 1 $ convolution layer. Interestingly, the CCT (BiLSTM) version can lead to  2.6\% improvement. However, the CCT( $1\times1$ Conv), CCT (Global Conv), CCT (Large Kernel ) and CCT (Stacked Conv) version is slightly smaller, which verifies that modeling global interaction to generate the parameters of CCT is reasonable and very essential.

\begin{figure}[t]
\begin{center}
   \includegraphics[width=\linewidth]{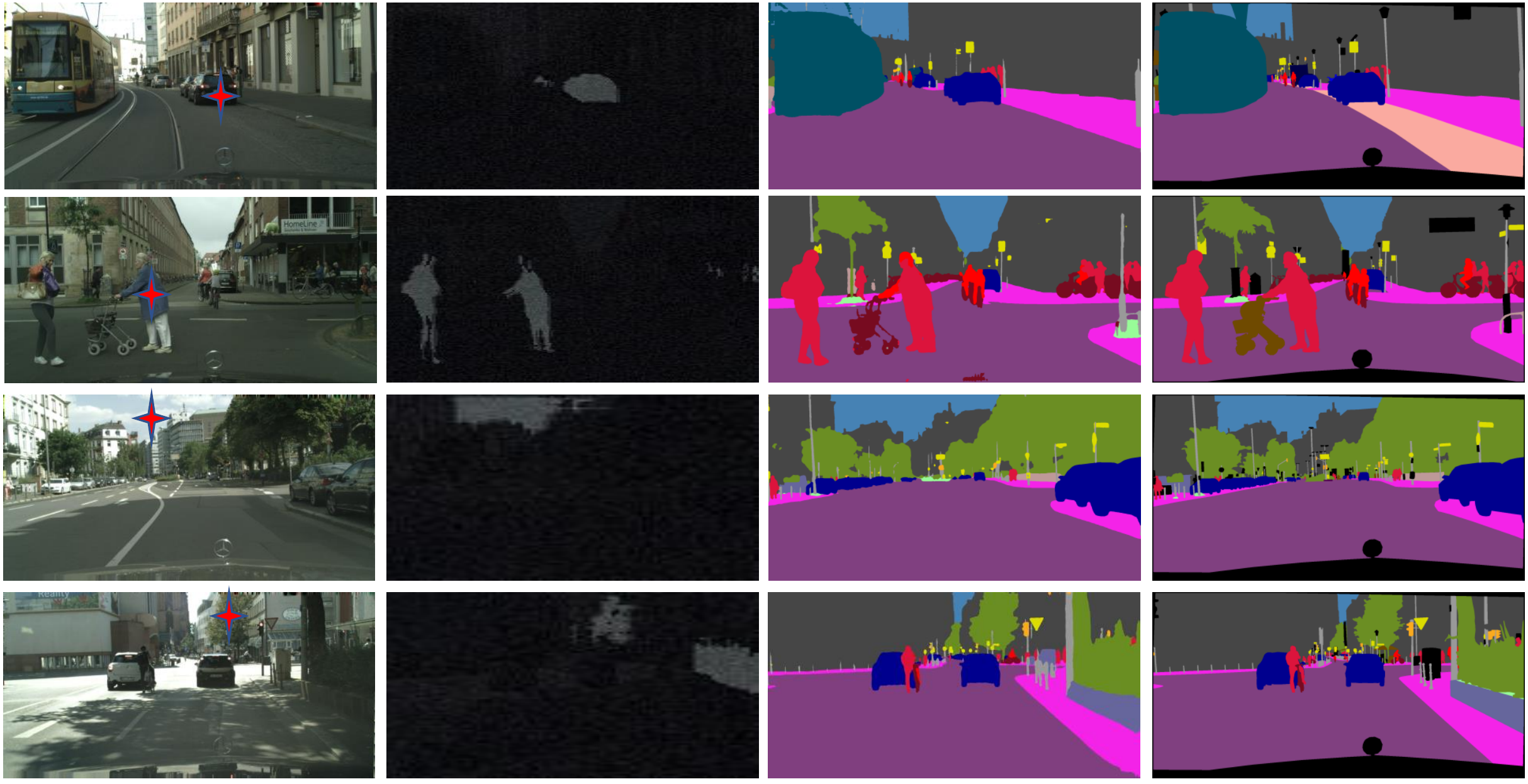}
\end{center}
   \caption{Visualization result of the category consensus transform on Cityscapes validation set. From left to right are: input image, the parameter maps of CCT, prediction and ground truth.}
\label{fig:fig4}
\vspace{-15pt}
\end{figure}

\subsubsection{CCT unit vs. Non-local unit}
Tab.~\ref{table:tab4} compares our CCT unit with non-local unit \cite{wang2018non} (denoted as ``+NL"). The non-local unit can generate attention masks for each position by considering the pair-wise feature correlation, and compute the response at a position as a weighted sum of the features at all positions. The proposed CCT unit can achieve better performance than ``Res101 + NL ". Here we give some possible explanations: (1) For the non-local unit, although the response at any position is a weighted sum of the features at all positions, the weight parameters are computed by the feature which has the limited receptive field, and does not model complex global interaction. (2) In contrast to them, the proposed CCT unit employs row-wise and column-wise BiLSTM to scan the whole feature map, which has a global receptive field.
Therefore, the process of generating the parameters of CCT for each position model the interaction at the specific location with all other positions.

\subsubsection{Intergrating ICT and CCT unit}

Next, we conduct experiments with different settings in Tab.~\ref{table:tab5} to verify the effectiveness of consensus transforms. As shown in Tab.~\ref{table:tab5}, the consensus transform units can improve the performance remarkably. Compared with the baseline method,  employing the ICT unit can yield a result of 78.8\% in mean IoU, which has 3.8\% improvement.
Meanwhile, employing the CCT unit can individually bring 2.6\% improvement over the baseline. When we integrate the ICT and CCT units, the performance is further improved to 79.9\%, which has 5.0\% improvement over the baseline (79.9 vs. 74.9). These experiments show that the integration of both units can bring great benefit to scene parsing. In addition, we report the computation overhead of the proposed units in Tab..~\ref{table:tab5}, which is estimated for the input of $3 \times 512 \times 512$. The inference time of ICT and CCT increase by only $4.6\%$ and $16.5\%$ respectively

\begin{table}[h]
\begin{center}
\begin{tabular}{|p{2.5cm}p{2.2cm}<{\centering}p{2cm}<{\centering}|}
\hline
Method &  Backbone & mIoU (\%) \\
\hline\hline
DeepLab-v2 \cite{chen2016deeplab} $\ddagger$  &ResNet-101    &45.7\\
RefineNet \cite{lin2017refinenet}   &ResNet-152    &47.3\\
CCL \cite{Ding_2018_CVPR}           &ResNet-101    & 51.6 \\
EncNet \cite{zhang2018context}      &ResNet-101    & 51.7 \\
TKCN \cite{wu2018tree}              &ResNet-101      &51.8\\
\hline
FCN (baseline)                       &ResNet-101      & 44.1 \\
CFNet (ours)                                &ResNet-101      &52.4\\
\hline
\end{tabular}
\end{center}
\caption{Accuracy comparison of our method against other methods on PASCAL Context test set.  $\ddagger$ indicates using extra training data form COCO ( 100K images), while we only use the PASCAL Context training data. }
\label{table:tab7}
\end{table}
\begin{table}[ht]
\begin{center}
\begin{tabular}{|p{2.6cm}p{2.2cm}<{\centering}p{2cm}<{\centering}|}

\hline
Method &  Backbone & mIoU (\%) \\
\hline\hline
SegNet \cite{badrinarayanan2017segnet}  &VGG-16                 &  55.6 \\
CGNet \cite{wu2018cgnet}                & -                      & 65.6 \\
G-FRNet \cite{amirul2017gated}          &VGG-16                 &  68.0\\
BiSeNet \cite{yu2018bisenet}            &ResNet-18              &  68.7 \\
DenseDecoder \cite{bilinski2018dense}   &ResNeXt-101            &  70.9 \\
\hline
FCN (baseline)                       &ResNet-101       &  67.5   \\
CFNet (ours)                               &ResNet-101      &  71.6 \\
\hline
\end{tabular}

\end{center}
\caption{Accuracy comparison of our method against other methods on CamVid test set. }
\label{table:tab8}
\vspace{-15pt}
\end{table}
\subsubsection{Visualization an Analysis}
In this subsection, we give some qualitative proof of the proposed consensus transforms. We visualize the parameters of the CCT unit.
According to Eqn.~\eqref{eq10}, each position has $H \times W$ parameters, and each transformed feature is a linear combination of all input features. As shown in Fig.~\ref{fig:fig4}, we choose one point (marked as \textcolor[rgb]{1,0,0}{+}) in each image and visualize their parameters  in the second column. We can observe that the category-level consensus transform focuses on aggregating features of the same semantic category. For example, for the point in the first row, its parameter map focus on the position which belongs to the ``car'' category, which demonstrates that our proposed CCT unit is very helpful for learning category-level consensus features.

\subsubsection{Comparison with state-of-the-arts}
We further compare the proposed methods with existing methods on the Cityscapes test set. Specifically, we just only use fine annotated data to train our CFNet and submit our test results to the official evaluation server. Performance is shown the Tab.~\ref{table:tab6}, the proposed CFNet can achieve 81.3\% in mean IoU. Note that the performance is slightly lower than the very recent CCNet \cite{Huang_2019_ICCV}, which adopts the online hard example mining (OHEM). 

\subsection{Experiments on PASCAL Context}
We conduct experiments on the PASCAL Context dataset to further verify the effectiveness of our approach.  The crop-size is set to $521 \times 521$, and training times are set to 30 epochs. Other training and testing setting are the same as that on the Cityscapes dataset. The quantitative results of this dataset are shown in Tab.~\ref{table:tab7}. The proposed CFNet achieve 52.4\% in mean IoU, substantially brings an 8.3\% improvement over the baseline (52.4\% vs. 44.1\%). Among existing works, most of them use multiple scale feature learning or employ 
\begin{table}[t]
\begin{center}

\begin{tabular}{|p{2.5cm}p{2.2cm}<{\centering}p{2cm}<{\centering}|}
\hline
Method &  Backbone & mIoU (\%) \\
\hline\hline
DeepLab-v2 \cite{chen2016deeplab}   &ResNet-101    &26.9\\
DAG-RNN  \cite{shuai2018scene}      &VGG-16        &30.4\\
RefineNet \cite{lin2017refinenet}   &ResNet-101    &33.6\\
CCL \cite{Ding_2018_CVPR}           &ResNet-101    & 35.7 \\
\hline
FCN (baseline)                       &ResNet-101      &30.2   \\
CFNet (ours)                               &ResNet-101      & 36.6     \\
\hline
\end{tabular}
\end{center}
\caption{Accuracy comparison of our method against other methods on COCO Stuff test set. }
\label{table:tab9}
\vspace{-15pt}
\end{table}
\noindent context modules to improve performance. In contrast to them, we introduce the two consensus transform units to learn the instance-level consensus and category-level consensus features. and the proposed approach can achieve better parsing results.

\subsection{Experiments on CamVid}
We conduct experiments on the CamVid dataset to further verify the effectiveness of our approach. To make our experimental setting comparable to previous works \cite{kundu2016feature, badrinarayanan2017segnet, amirul2017gated, bilinski2018dense}, we downsample the images in the dataset by a factor of 2. The base learning rate is set to 0.025, crop-size is set to $480 \times 360$, and training times are set to 100 epochs. The quantitative results of this dataset are shown in Tab.~\ref{table:tab8}. The baseline method achieves 67.5\%. The proposed CFNet achieve 71.6\%, which outperforms previous state-of-the-art method DenseDecoder \cite{bilinski2018dense}, which introduces dense decoder shortcut connections for fuse semantic feature maps form all previous decoder levels.

\subsection{Experiments on COCO Stuff}
Finally, we further run our method on the COCO Stuff dataset for demonstrating the generality of the proposed CFNet. 
The crop-size is set to $521 \times 641$, and training times are set to 25 epochs. 
The experiment results as shown in Tab.~\ref{table:tab9}. The baseline achieves 30.2\% in mean IoU. Our method achieves 36.6\% mean IoU, which outperforms the previous state-of-the-art method CCL \cite{Ding_2018_CVPR}.

\section{Conclusion}

In this work, we propose the Instance Consensus Transforms and Category Consensus Transform units to learn the instance-level and category-level consensus features, which is desired for scene parsing. Based on the proposed two units, we develop a novel framework called Consensus Feature Network (CFNet). The ablation experiments demonstrate that the proposed approach can effectively learn the pixel-wise consensus features, and obtain consistent parsing results. Furthermore, we show the advantages of CFNet with state-of-the-art performance on four benchmarks including Cityscapes, PASCAL Context, CamVid, and COCO Stuff.


{\small
\bibliographystyle{ieee}
\bibliography{egbib}

\begin{thebibliography}{10}\itemsep=-1pt

\bibitem{amirul2017gated}
M.~Amirul~Islam, M.~Rochan, N.~D. Bruce, and Y.~Wang.
\newblock Gated feedback refinement network for dense image labeling.
\newblock In {\em CVPR}, 2017.

\bibitem{badrinarayanan2017segnet}
V.~Badrinarayanan, A.~Kendall, and R.~Cipolla.
\newblock Segnet: A deep convolutional encoder-decoder architecture for image
  segmentation.
\newblock {\em TPAMI}, 2017.

\bibitem{belongie2002shape}
S.~Belongie, J.~Malik, and J.~Puzicha.
\newblock Shape matching and object recognition using shape contexts.
\newblock Technical report, 2002.

\bibitem{bian2017gms}
J.~Bian, W.-Y. Lin, Y.~Matsushita, S.-K. Yeung, T.-D. Nguyen, and M.-M. Cheng.
\newblock Gms: grid-based motion statistics for fast, ultra-robust feature
  correspondence.
\newblock In {\em CVPR}, 2017.

\bibitem{bilinski2018dense}
P.~Bilinski and V.~Prisacariu.
\newblock Dense decoder shortcut connections for single-pass semantic
  segmentation.
\newblock In {\em CVPR}, 2018.

\bibitem{brostow2008segmentation}
G.~J. Brostow, J.~Shotton, J.~Fauqueur, and R.~Cipolla.
\newblock Segmentation and recognition using structure from motion point
  clouds.
\newblock In {\em ECCV}, 2008.

\bibitem{Caesar_2018_CVPR}
H.~Caesar, J.~Uijlings, and V.~Ferrari.
\newblock Coco-stuff: Thing and stuff classes in context.
\newblock In {\em CVPR}, 2018.

\bibitem{chang1997fast}
S.-H. Chang, F.-H. Cheng, W.-H. Hsu, and G.-Z. Wu.
\newblock Fast algorithm for point pattern matching: invariant to translations,
  rotations and scale changes.
\newblock {\em Pattern recognition}, 1997.

\bibitem{chen2016deeplab}
L.-C. Chen, G.~Papandreou, I.~Kokkinos, K.~Murphy, and A.~L. Yuille.
\newblock Deeplab: Semantic image segmentation with deep convolutional nets,
  atrous convolution, and fully connected crfs.
\newblock {\em arXiv preprint arXiv:1606.00915}, 2016.

\bibitem{chen2017rethinking}
L.-C. Chen, G.~Papandreou, F.~Schroff, and H.~Adam.
\newblock Rethinking atrous convolution for semantic image segmentation.
\newblock 2017.

\bibitem{chen2019graph}
Y.~Chen, M.~Rohrbach, Z.~Yan, Y.~Shuicheng, J.~Feng, and Y.~Kalantidis.
\newblock Graph-based global reasoning networks.
\newblock In {\em Proceedings of the IEEE Conference on Computer Vision and
  Pattern Recognition}, pages 433--442, 2019.

\bibitem{Cheng_2019_ICCV}
B.~Cheng, L.-C. Chen, Y.~Wei, Y.~Zhu, Z.~Huang, J.~Xiong, T.~S. Huang, W.-M.
  Hwu, and H.~Shi.
\newblock Spgnet: Semantic prediction guidance for scene parsing.
\newblock In {\em The IEEE International Conference on Computer Vision (ICCV)},
  October 2019.

\bibitem{cordts2016cityscapes}
M.~Cordts, M.~Omran, S.~Ramos, T.~Rehfeld, M.~Enzweiler, R.~Benenson,
  U.~Franke, S.~Roth, and B.~Schiele.
\newblock The cityscapes dataset for semantic urban scene understanding.
\newblock In {\em CVPR}, 2016.

\bibitem{Ding_2018_CVPR}
H.~Ding, X.~Jiang, B.~Shuai, A.~Qun~Liu, and G.~Wang.
\newblock Context contrasted feature and gated multi-scale aggregation for
  scene segmentation.
\newblock In {\em CVPR}, 2018.

\bibitem{fu2018dual}
J.~Fu, J.~Liu, H.~Tian, Z.~Fang, and H.~Lu.
\newblock Dual attention network for scene segmentation.
\newblock {\em arXiv preprint arXiv:1809.02983}, 2018.

\bibitem{Fu_2019_ICCV}
J.~Fu, J.~Liu, Y.~Wang, Y.~Li, Y.~Bao, J.~Tang, and H.~Lu.
\newblock Adaptive context network for scene parsing.
\newblock In {\em The IEEE International Conference on Computer Vision (ICCV)},
  October 2019.

\bibitem{harley2015learning}
A.~W. Harley, K.~G. Derpanis, and I.~Kokkinos.
\newblock Learning dense convolutional embeddings for semantic segmentation.
\newblock {\em arXiv preprint arXiv:1511.04377}, 2015.

\bibitem{harley2017segmentation}
A.~W. Harley, K.~G. Derpanis, and I.~Kokkinos.
\newblock Segmentation-aware convolutional networks using local attention
  masks.
\newblock In {\em ICCV}, 2017.

\bibitem{he2016deep}
K.~He, X.~Zhang, S.~Ren, and J.~Sun.
\newblock Deep residual learning for image recognition.
\newblock In {\em CVPR}, 2016.

\bibitem{hou2018self}
Q.~Hou, P.~Jiang, Y.~Wei, and M.-M. Cheng.
\newblock Self-erasing network for integral object attention.
\newblock In {\em NIPS}, 2018.

\bibitem{huang2017densely}
G.~Huang, Z.~Liu, L.~Van Der~Maaten, and K.~Q. Weinberger.
\newblock Densely connected convolutional networks.
\newblock In {\em CVPR}, 2017.

\bibitem{Huang_2019_ICCV}
Z.~Huang, X.~Wang, L.~Huang, C.~Huang, Y.~Wei, and W.~Liu.
\newblock Ccnet: Criss-cross attention for semantic segmentation.
\newblock In {\em The IEEE International Conference on Computer Vision (ICCV)},
  October 2019.

\bibitem{Ke_2018_ECCV}
T.-W. Ke, J.-J. Hwang, Z.~Liu, and S.~X. Yu.
\newblock Adaptive affinity fields for semantic segmentation.
\newblock In {\em ECCV}, 2018.

\bibitem{kundu2016feature}
A.~Kundu, V.~Vineet, and V.~Koltun.
\newblock Feature space optimization for semantic video segmentation.
\newblock In {\em CVPR}, 2016.

\bibitem{Li_2019_ICCV}
X.~Li, Z.~Zhong, J.~Wu, Y.~Yang, Z.~Lin, and H.~Liu.
\newblock Expectation-maximization attention networks for semantic
  segmentation.
\newblock In {\em The IEEE International Conference on Computer Vision (ICCV)},
  October 2019.

\bibitem{lin2017refinenet}
G.~Lin, A.~Milan, C.~Shen, and I.~D. Reid.
\newblock Refinenet: Multi-path refinement networks for high-resolution
  semantic segmentation.
\newblock In {\em CVPR}, 2017.

\bibitem{lin2014microsoft}
T.-Y. Lin, M.~Maire, S.~Belongie, J.~Hays, P.~Perona, D.~Ramanan,
  P.~Doll{\'a}r, and C.~L. Zitnick.
\newblock Microsoft coco: Common objects in context.
\newblock In {\em ECCV}, 2014.

\bibitem{liu2018picanet}
N.~Liu, J.~Han, and M.-H. Yang.
\newblock Picanet: Learning pixel-wise contextual attention for saliency
  detection.
\newblock In {\em Proceedings of the IEEE Conference on Computer Vision and
  Pattern Recognition}, pages 3089--3098, 2018.

\bibitem{liu2015parsenet}
W.~Liu, A.~Rabinovich, and A.~C. Berg.
\newblock Parsenet: Looking wider to see better.
\newblock {\em arXiv preprint arXiv:1506.04579}, 2015.

\bibitem{long2015fully}
J.~Long, E.~Shelhamer, and T.~Darrell.
\newblock Fully convolutional networks for semantic segmentation.
\newblock In {\em CVPR}, 2015.

\bibitem{maaten2008visualizing}
L.~v.~d. Maaten and G.~Hinton.
\newblock Visualizing data using t-sne.
\newblock {\em Journal of machine learning research}, 2008.

\bibitem{mottaghi2014role}
R.~Mottaghi, X.~Chen, X.~Liu, N.-G. Cho, S.-W. Lee, S.~Fidler, R.~Urtasun, and
  A.~Yuille.
\newblock The role of context for object detection and semantic segmentation in
  the wild.
\newblock In {\em CVPR}, 2014.

\bibitem{peng2017large}
C.~Peng, X.~Zhang, G.~Yu, G.~Luo, and J.~Sun.
\newblock Large kernel matters--improve semantic segmentation by global
  convolutional network.
\newblock {\em arXiv preprint arXiv:1703.02719}, 2017.

\bibitem{rocco2018end}
I.~Rocco, R.~Arandjelovi{\'c}, and J.~Sivic.
\newblock End-to-end weakly-supervised semantic alignment.
\newblock In {\em CVPR}, 2018.

\bibitem{Rocco18b}
I.~Rocco, M.~Cimpoi, R.~Arandjelovi\'c, A.~Torii, T.~Pajdla, and J.~Sivic.
\newblock Neighbourhood consensus networks.
\newblock In {\em Proceedings of the 32nd Conference on Neural Information
  Processing Systems}, 2018.

\bibitem{rota2018place}
S.~Rota~Bul{\`o}, L.~Porzi, and P.~Kontschieder.
\newblock In-place activated batchnorm for memory-optimized training of dnns.
\newblock In {\em CVPR}, 2018.

\bibitem{schalkoff1989digital}
R.~J. Schalkoff.
\newblock {\em Digital image processing and computer vision}, volume 286.
\newblock Wiley New York, 1989.

\bibitem{schmid1997local}
C.~Schmid and R.~Mohr.
\newblock Local grayvalue invariants for image retrieval.
\newblock {\em TPAMI}, 1997.

\bibitem{shuai2018scene}
B.~Shuai, Z.~Zuo, B.~Wang, and G.~Wang.
\newblock Scene segmentation with dag-recurrent neural networks.
\newblock {\em TPAMI}, 2018.

\bibitem{simonyan2014very}
K.~Simonyan and A.~Zisserman.
\newblock Very deep convolutional networks for large-scale image recognition.
\newblock {\em arXiv preprint arXiv:1409.1556}, 2014.

\bibitem{stockman1982matching}
G.~Stockman, S.~Kopstein, and S.~Benett.
\newblock Matching images to models for registration and object detection via
  clustering.
\newblock {\em TPAMI}, 1982.

\bibitem{tomasi1998bilateral}
C.~Tomasi and R.~Manduchi.
\newblock Bilateral filtering for gray and color images.
\newblock In {\em null}. IEEE, 1998.

\bibitem{trulls2013dense}
E.~Trulls, I.~Kokkinos, A.~Sanfeliu, and F.~Moreno-Noguer.
\newblock Dense segmentation-aware descriptors.
\newblock In {\em CVPR}, 2013.

\bibitem{tu2005image}
Z.~Tu, X.~Chen, A.~L. Yuille, and S.-C. Zhu.
\newblock Image parsing: Unifying segmentation, detection, and recognition.
\newblock {\em IJCV}, 2005.

\bibitem{visin2015renet}
F.~Visin, K.~Kastner, K.~Cho, M.~Matteucci, A.~Courville, and Y.~Bengio.
\newblock Renet: A recurrent neural network based alternative to convolutional
  networks.
\newblock {\em arXiv preprint arXiv:1505.00393}, 2015.

\bibitem{wang2018non}
X.~Wang, R.~Girshick, A.~Gupta, and K.~He.
\newblock Non-local neural networks.
\newblock In {\em CVPR}, 2018.

\bibitem{wu2018tree}
T.~Wu, S.~Tang, R.~Zhang, and J.~Li.
\newblock Tree-structured kronecker convolutional networks for semantic
  segmentation.
\newblock {\em arXiv preprint arXiv:1812.04945}, 2018.

\bibitem{wu2018cgnet}
T.~Wu, S.~Tang, R.~Zhang, and Y.~Zhang.
\newblock Cgnet: A light-weight context guided network for semantic
  segmentation.
\newblock {\em arXiv preprint arXiv:1811.08201}, 2018.

\bibitem{yang2018denseaspp}
M.~Yang, K.~Yu, C.~Zhang, Z.~Li, and K.~Yang.
\newblock Denseaspp for semantic segmentation in street scenes.
\newblock In {\em CVPR}, 2018.

\bibitem{yu2018bisenet}
C.~Yu, J.~Wang, C.~Peng, C.~Gao, G.~Yu, and N.~Sang.
\newblock Bisenet: Bilateral segmentation network for real-time semantic
  segmentation.
\newblock 2018.

\bibitem{Yu_2018_CVPR}
C.~Yu, J.~Wang, C.~Peng, C.~Gao, G.~Yu, and N.~Sang.
\newblock Learning a discriminative feature network for semantic segmentation.
\newblock In {\em CVPR}, 2018.

\bibitem{yuan2018ocnet}
Y.~Yuan and J.~Wang.
\newblock Ocnet: Object context network for scene parsing.
\newblock {\em arXiv preprint arXiv:1809.00916}, 2018.

\bibitem{zeiler2014visualizing}
M.~D. Zeiler and R.~Fergus.
\newblock Visualizing and understanding convolutional networks.
\newblock In {\em European conference on computer vision}, pages 818--833.
  Springer, 2014.

\bibitem{zhang2018context}
H.~Zhang, K.~Dana, J.~Shi, Z.~Zhang, X.~Wang, A.~Tyagi, and A.~Agrawal.
\newblock Context encoding for semantic segmentation.
\newblock In {\em CVPR}, 2018.

\bibitem{zhang2017global}
R.~Zhang, S.~Tang, M.~Lin, J.~Li, and S.~Yan.
\newblock Global-residual and local-boundary refinement networks for rectifying
  scene parsing predictions.
\newblock In {\em IJCAI}, 2017.

\bibitem{zhang2017scale}
R.~Zhang, S.~Tang, Y.~Zhang, J.~Li, and S.~Yan.
\newblock Scale-adaptive convolutions for scene parsing.
\newblock In {\em ICCV}, 2017.

\bibitem{zhang2018adversarial}
X.~Zhang, Y.~Wei, J.~Feng, Y.~Yang, and T.~S. Huang.
\newblock Adversarial complementary learning for weakly supervised object
  localization.
\newblock In {\em CVPR}, pages 1325--1334, 2018.

\bibitem{zhang1995robust}
Z.~Zhang, R.~Deriche, O.~Faugeras, and Q.-T. Luong.
\newblock A robust technique for matching two uncalibrated images through the
  recovery of the unknown epipolar geometry.
\newblock {\em Artificial intelligence}, 1995.

\bibitem{Zhao_2017_CVPR}
H.~Zhao, J.~Shi, X.~Qi, X.~Wang, and J.~Jia.
\newblock Pyramid scene parsing network.
\newblock In {\em CVPR}, 2017.

\bibitem{Zhao_2018_ECCV}
H.~Zhao, Y.~Zhang, S.~Liu, J.~Shi, C.~Change~Loy, D.~Lin, and J.~Jia.
\newblock Psanet: Point-wise spatial attention network for scene parsing.
\newblock In {\em ECCV}, 2018.

\bibitem{Zhu_2019_ICCV}
Z.~Zhu, M.~Xu, S.~Bai, T.~Huang, and X.~Bai.
\newblock Asymmetric non-local neural networks for semantic segmentation.
\newblock In {\em The IEEE International Conference on Computer Vision (ICCV)},
  October 2019.

\end{thebibliography}
}

\clearpage
\newpage
\onecolumn 
 \appendix
\appendixpage
\addappheadtotoc

\section{Qualitative Results}
We give more qualitative results of the baseline model and our proposed approach on Cityscapes \cite{cordts2016cityscapes}, PASCAL Context \cite{mottaghi2014role}, CamVid \cite{brostow2008segmentation} and COCO Stuff \cite{Caesar_2018_CVPR}.

\begin{figure*}[ht]
\begin{center}
   \includegraphics[width=0.9\linewidth]{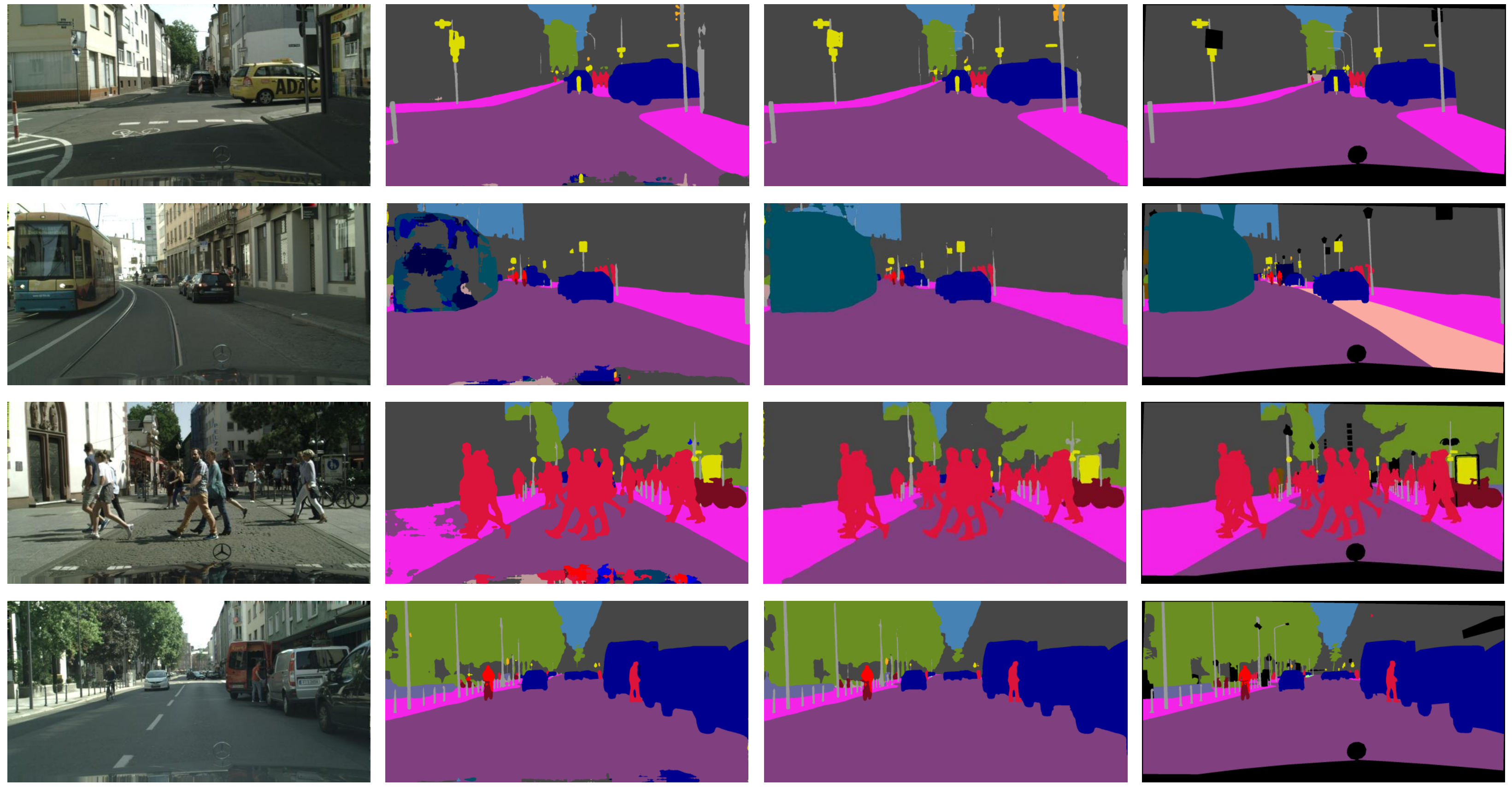}
\end{center}
   \caption{ Result illustration of the proposed CFNet on Cityscapes validation set. From left to right are: input image, prediction of the
baseline, prediction of the proposed CFNet, and ground truth. (Best viewed in color) }
\label{fig:fig2}
\end{figure*}

\begin{figure*}[t]
\begin{center}
   \includegraphics[width=0.83\linewidth]{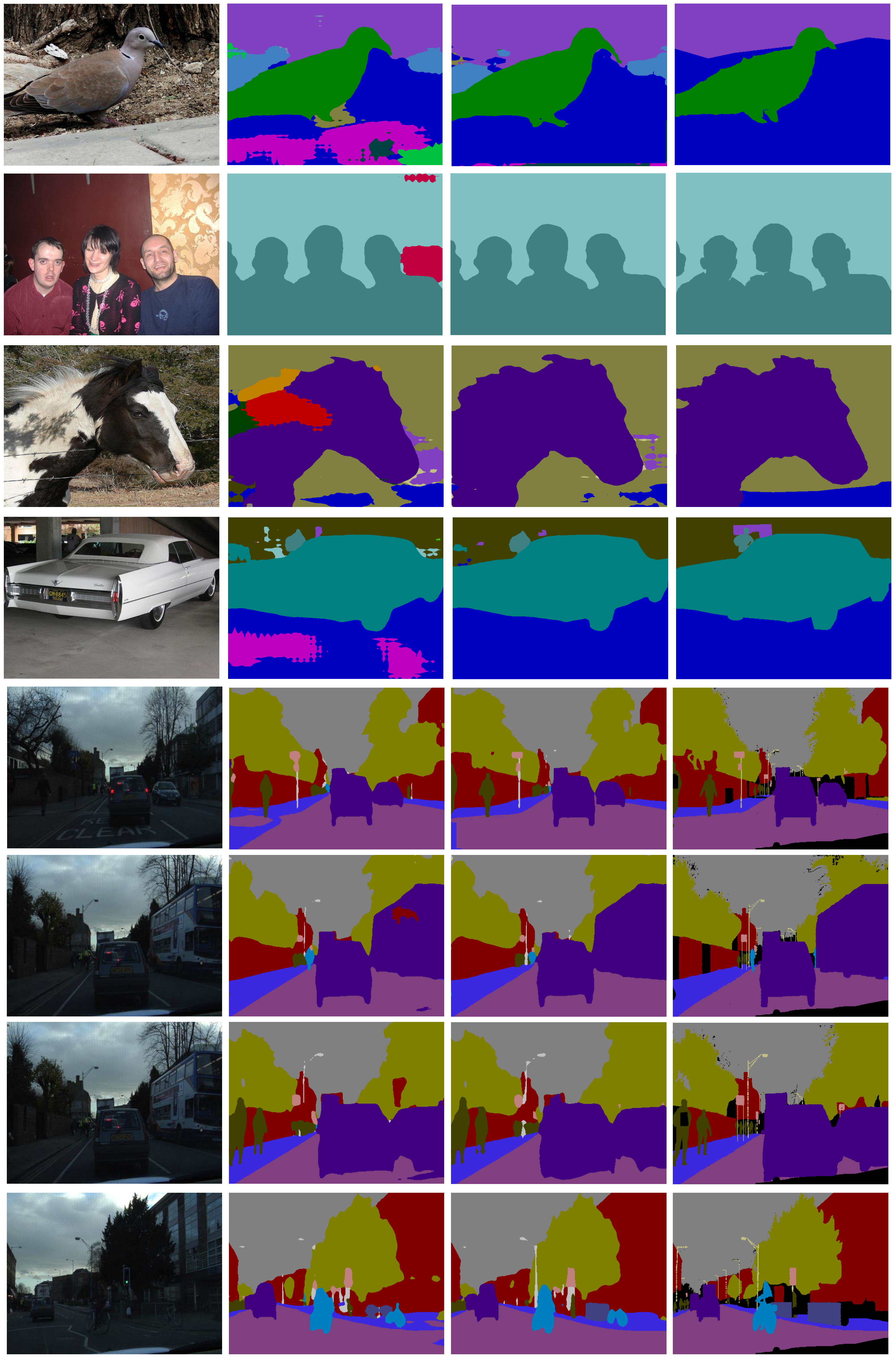}
\end{center}
   \caption{The first four rows show four examples from PASCAL Context test set; the last four rows illustrates 4 examples from CamVid
test set. From left to right are: input image, prediction of the baseline, prediction of the proposed CFNet, and ground truth. (Best viewed
in color)}
\label{fig:fig2}
\vspace{-15pt}
\end{figure*}

\begin{figure*}
\begin{center}
   \includegraphics[width=0.9\linewidth]{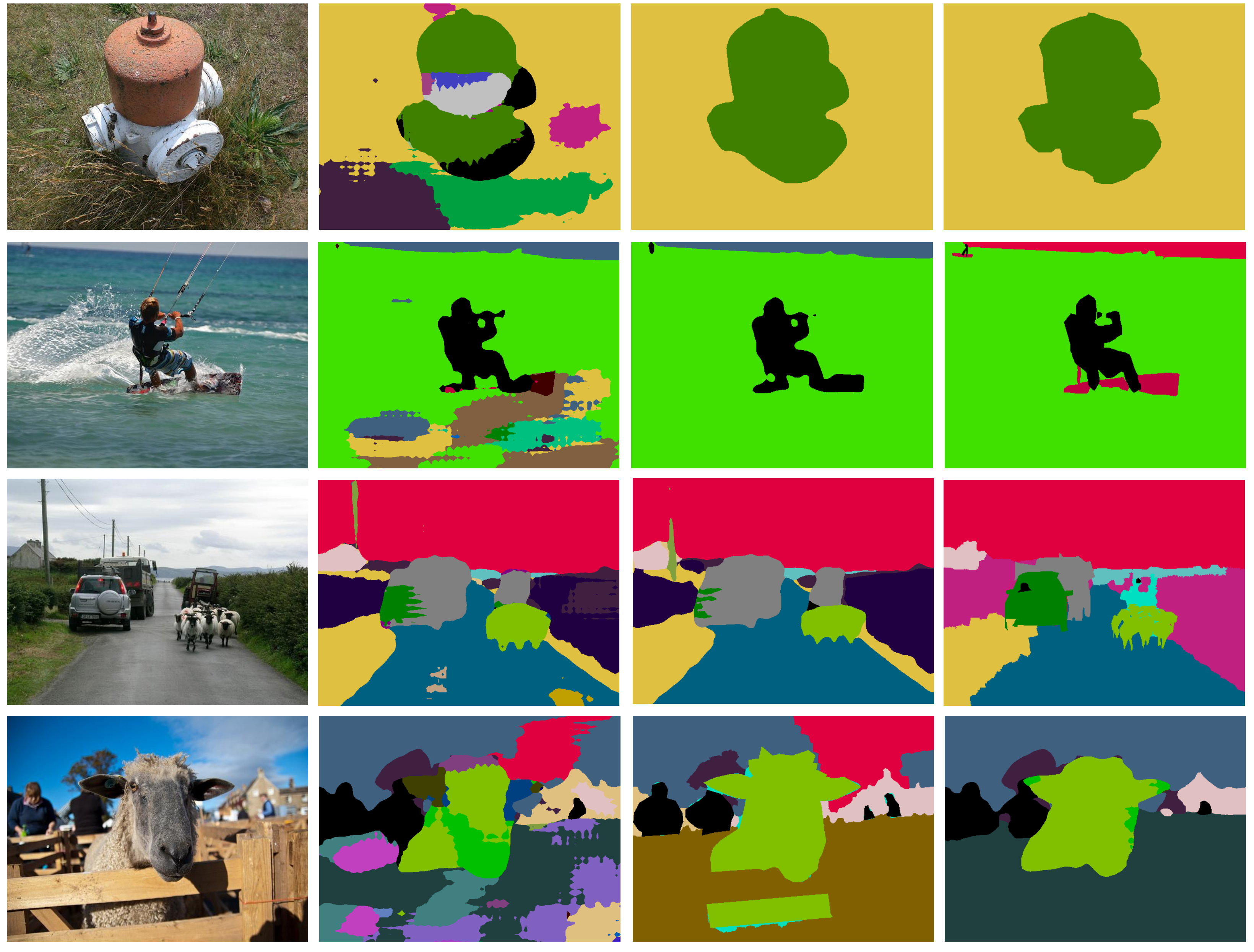}
\end{center}
   \caption{ Result illustration of the proposed CFNet on COCO Stuff validation set. From left to right are: input image, prediction of the
baseline, prediction of the proposed CFNet, and ground truth. (Best viewed in color) }
\label{fig:fig2}
\end{figure*}
\end{document}